\documentclass{article}

     \usepackage[final]{neurips_2019}

\usepackage[utf8]{inputenc} 
\usepackage[T1]{fontenc}    
\usepackage{hyperref}       
\usepackage{url}            
\usepackage{booktabs}       
\usepackage{amsfonts}       
\usepackage{nicefrac}       
\usepackage{microtype}      
\usepackage{natbib}
\usepackage{graphicx}
\usepackage{subfig}
\usepackage{multirow}
\usepackage{multicol}
\usepackage{amsmath}
\title{Bayesian Weight Priors for Learning Identity Relations}
\title{Weight Priors for Learning Identity Relations}

\author{
    Radha Kopparti \\
  Department of Computer Science \\
  City, University of London\\
  London, United Kingdom \\
  \texttt{radha.kopparti@city.ac.uk} \\
  \And
  Tillman Weyde \\
  Department of Computer Science \\
  City, University of London\\
  London, United Kigndom \\
  \texttt{t.e.weyde@city.ac.uk} \\
 }
 
\begin{document}

\maketitle

\begin{abstract}
Learning abstract and systematic relations has been an open issue in neural network learning for over 30 years. 
It has been shown recently that neural networks do not learn relations based on identity and are unable to generalize well to unseen data. 
The Relation Based Pattern (RBP) approach has been proposed as a solution for this problem. 
In this work, we extend RBP by realizing it as a Bayesian prior on network weights to model the identity relations. 
This weight prior leads to a modified regularization term in otherwise standard network learning.
In our experiments, we show that the Bayesian weight priors lead to perfect generalization when learning identity based relations and do not impede general neural network learning. 
We believe that the approach of creating an inductive bias with weight priors can be extended easily to other forms of relations and will be beneficial for many other learning tasks.
\end{abstract}

\section{Introduction}\label{sec:intro}
Humans, including infants. are very effective at learning patterns of relations based on identity from sensory input and systematically applying them to new stimuli, even after a very brief exposure, 
while current neural networks in their standard form fail to detect these relations in unseen data \citep{GaryMarcus1999}.
It has become evident that there are still relevant limitations to systematic generalization in current neural network architectures \citep{lake2018generalization,GaryMarcus2018}. 
Neural networks are seen to be good at memorizing the numerical patterns seen in the training set but often not to extrapolate this representation outside the training set \citep{liska-et-al-2018-memorize}.
It was found that standard neural networks do not seem to learn identity relations, i.e. the equality of two vectors  \citep{weyde_kopparti_2018}, which are fundamental for many higher level tasks. 

A well known study in this direction was conducted by \cite{GaryMarcus1999}, where a recurrent neural network failed to distinguish abstract patterns, based on equality or identity relations between the input stimuli, although seven-month-old infants showed the ability to distinguish them after a few minutes of exposure. 
This was followed by an lively exchange on rule learning by neural networks and in human language acquisition in general, where results by \cite{Elman1999, Altmann2} and \cite{Shultz1} could not be reproduced by \cite{vilcu2001generalization,vilcu2005two}, and \cite{shultz2006neural} disputed claims by \cite{vilcu2005two}.
Other approaches, such as those by \cite{Shastri-1999-spatiotemporal} and \cite{Dominey,Alhama}, use specialized network architectures or different problem formulations or evaluation methods. 

Recently, the Relation Based Patterns (RBP) approach has been introduced as a way to create a suitable inductive bias for the problem of learning identity relations on binary vectors \cite{weyde_kopparti_2018,weyde_kopparti_2019}. 
The task is to classify whether two halves of the input vectors are equal. i.e. $u_i = u_{n+i} \forall i \in \{1, \dots , n\}$ for an input vector with $2n$ dimensions. 
The RBP model introduced in \cite{weyde_kopparti_2018,weyde_kopparti_2019} is based on the comparison of input neurons that correspond to each other in a relation, e.g. the corresponding dimensions in a pair of binary vectors. 
For the comparison they introduce Differentiator-Rectifier (DR) units, which calculate the absolute difference of two inputs: $f(x,y) = |x-y|$. 
For each dimension of the input vectors, a DR unit is introduced. 
This simplifies the learning problem because the summation of activations of the DR units is sufficient for generalisable identity detection. 
The results show no restriction on the learning of other tasks in practice. 

There are different ways of integrating DR units into neural networks in RBP: \textit{Early} and  \textit{Mid} Fusion.
In \emph{Early Fusion}, DR units are concatenated to input units, and in \emph{Mid Fusion} they are concatenated to the hidden layer. 
In both cases, the existing input and hidden units are unchanged. 

Adding units with hard-wired connections and unusual activation function limits the flexibility of the RBP approach. 
In this work, we introduce a modified RBP structure that can be formulated as a Bayesian prior to model the weight structure of Mid Fusion RBP in a standard feed forward network setting.

\section{A Bayesian Approach to Relation Based Patterns}
In the weight prior approach, we replace each DR units with two standard neurons and model the fixed weights with a default weight matrix $D$. 
This matrix contains the weights that enable a dimension-wise comparison of the inputs. 

\begin{figure}[h]

        \begin{tabular}{p{6cm}c} 
        a) $\vcenter{\hbox{{\includegraphics[width=5cm]{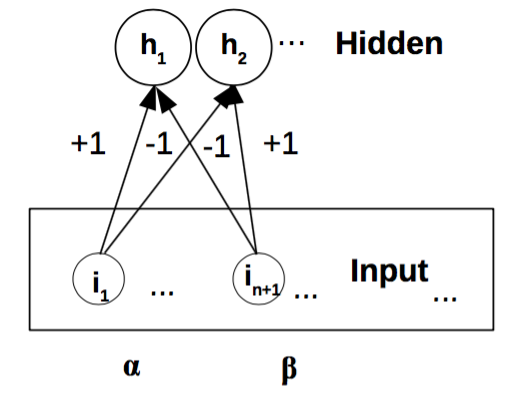} }}}$
            &
            b)
              {$\displaystyle
                D = {\text{\boldmath$
                \begin{pmatrix} 
                     +1  & 0 & 0 & -1 & 0 & 0 \\
                    -1  & 0 & 0 & +1 & 0 & 0 \\
                     0  & +1 & 0 & 0 & -1 & 0 \\
                     0  & -1 & 0 & 0 & +1 & 0 \\
                     0  & 0 & +1 & 0 & 0 & -1 \\
                     0  & 0 & -1 & 0 & 0 & +1 \\
                     0  & 0 & 0 & 0 & 0 & 0 \\
                &\vdots & & &\vdots & \\
                \end{pmatrix}$}}
            $}
        \end{tabular}
        \caption{a) 
        For each input dimension there are two hidden layer neurons, e.g. $h_1,h_2$ for dimension $1$.
        $\alpha$ and $\beta$ indicate two input vectors of dimensionality $n$. 
        b) 
        The default weight matrix $D$ for vector dim \textit{n}=3. 
        Each row corresponds to the incoming weights of a hidden neuron, e.g. the first row to $h_1$ in figure \ref{fig:rbp4} a). 
        If there are more hidden neurons than pairs of input vectors, the additional rows contain only zeros.}
        \label{fig:rbp4}
    \end{figure}

A diagram of the structure is given in Figure~\ref{fig:rbp4}(a) where $\alpha$ and $\beta$ are the two input values being compared. 
A small example of this default matrix is shown in Figure~\ref{fig:rbp4}b).
The incoming connections from two corresponding inputs to a neuron in the hidden layer to be compared have values of 1 and $-1$. 
For the same pair we use another hidden neuron with inverted signs of the weights, as in rows 1 and 2.
With a ReLU activation function, this means that the value of one of the hidden neurons will be positive, if one input neurons have different activations. 
We therefore need at least $n$ hidden units in the first hidden layer for a comparison of n two $n/2$-dimensional vectors. 
All other incoming connection weights are set to 0, including the bias. 
This ensures that in all cases where corresponding inputs are not equal, there will be a positive activation in one of the neurons.

The matrix $D$ is then used to define default values of the network weights, i.e. we impose a loss based on the difference between $D$ and the actual weight matrix $W$, that connects the input neurons to the first hidden layer. 
This loss term $l_{RBP_{1/2}}$ for L1 or L2 loss are defined as:
\begin{equation}
    l_{RBP_1} =  \sum_{i=1}^k  |W_k - D_k|, \hspace{1cm} l_{RBP_2} =  \sum_{i=1}^k  (W_k - D_k)^2,
\end{equation}
where $k$ is the number of elements in $W$. 
These loss functions correspond to Bayesian priors on the weights with the mean defined by the values of $D$. 
The $L2$ loss corresponds to a Gaussian and $L1$ loss to a Laplacian prior, such that backpropagation maximizes the posterior likelihood of the weights given the data \cite{williams1995bayesian}. 
The overall training loss $l_t$ is defined as 
\begin{equation}
    l_t = l_c  + \lambda \times l_{RBP}
\end{equation}
where $l_c$ is the cross entropy 
and $\lambda$ is the regularization parameter, corresponding to the inverse of the variance of the prior, effectively regulating the strength of the RBP regularization.
We call these methods of embedding the RBP into a standard network ERBP L1 and ERBP L2 respectively. 

\section{Generalizing and Learning of Identity Relations} 

For the task of learning identity relations, we generate synthetic data and use a standard feed-forward neural network. 
The input vector is binary and the target values are $[0,1]$ for unequal and $[1,0]$ for equal vector halves as described in section \ref{sec:intro}. 

We use a grid search over hyper-parameters: the number of epochs [10,20,30],
the number of neurons per hidden layer was varied as [10,20,30]. 
For the Bayesian weight prior, we varied the regularization parameter $\lambda$ with values [0.01,0.03,0.1,0.3,1,3,10,30].
We ran a total of 10 simulations using the SGD and Adam optimizers, training for 20 epochs. 
We used a single hidden layer and a batch size of 1 unless indicated otherwise. 
The networks have been implemented using the PyTorch library\footnote{\url{http://pytorch.org}}. 

The train/test split was set to 75\% / 25\% for all tasks. 
We tested with vector dimensionalities $n=3,10,30$. 
We generate all vectors with equal halves and take a random sample of all those with unequal halves to balance the classes. 
We downsample the size of the dataset to 1000 when it is greater.

\subsection{Identity Relations}
Identity is an abstract relation in the sense that it is independent of the actual values of the individual arguments, it just depends on their combined configuration.
In this task, pairs of vectors are presented to a feed-forward network and the task is to distinguish whether the two vectors are equal or not. 
We evaluated the performance of the network in different configurations on a held-out test set and the results are tabulated in Table~\ref{tab:exprs} for different vector dimensions.
As observed by \cite{weyde_kopparti_2018}, standard networks do not improve much over random guessing, while ERBP L1 and ERBP L2, as well as Mid Fusion, almost always achieve perfect generalizsation with sufficiently strong $\lambda$ (see next section for details).

\begin{table}[hbt]
\caption{
Test set classification accuracy (in \%) and standard deviation over 10 simulations (in brackets) using different models for Identity Learning (vector dimensions $n=3,10,30$). 
The networks were trained with the Adam optimizer for 20 epochs. 
}
\label{tab:exprs}
\centering
\begin{tabular}{lccccc}  
\toprule
Type & Standard & Early Fusion & Mid Fusion & ERBP L1 & ERBP L2 \\
\midrule
$n=3$ & 55 (1.91)  & 65 (1.34) & 100 (0.04)  & 100 (0.00) & 100 (0.00) \\
$n=10$ & 51 (1.67) & 65 (1.32) &100 (0.08) & 100 (0.04) & 100 (0.02) \\
$n=30$ & 50 (1.52) & 65 (1.27) & 100 (0.07) &100 (0.05) & 100 (0.04)\\
\bottomrule
\end{tabular}

\end{table}

\subsection{Parameter Variations}
We study the effect of several parameters: number of hidden layers, choice of optimizer, regularization factor and  weight initialization on identity relation learning tasks.

\textbf{Network Depth}:
We tested identity learning with deeper neural network models, using $h = {2,3,4,5}$ hidden layers. 
The results are tabulated in Table~\ref{tab:hid}, showing only minor improvements in the network performance for deeper networks. 
However, ERBP L1 and L2 generalization is consistent and independent of the network depth.

\begin{table}[htb]
\caption{Test set classification  accuracy (in \%) with standard deviation (in brackets) for identity learning ($n=3$) using deeper networks. The networks were trained with the Adam optimizer for 20 epochs. }
\label{tab:hid}
\centering
\begin{tabular}{cccccc}  
\toprule
Hidden  & No    & Early   & Mid  & ERBP L1 & ERBP L2   \\ 
layers  &  RBP   & Fusion  & Fusion & & \\
\midrule
h = 2  & 55 (1.65) & 65 (1.26)  & 100 (0.02) & 100 (0.00) & 100 (0.00) \\
h = 3 & 55 (1.67) & 67 (1.14) & 100 (0.03) & 100 (0.00) & 100 (0.00) \\ 
h = 4 & 58 (1.63) & 68 (1.25) & 100 (0.02) & 100 (0.00) & 100 (0.00)\\ 
h = 5 & 59 (1.68) & 72 (1.23) & 100 (0.02)  & 100 (0.00) & 100 (0.00)\\

\bottomrule
\end{tabular}

\end{table}

\begin{table}[htb]
\caption{Accuracy (in \%) and standard deviation of identity learning ($n=3$) using Adam and SGD optimizer for ERBP L1 and L2. SGD can also lead to 100\% accuracy on the test set, but needs higher $\lambda$ values.}
\label{tab:opt1}
\centering
\begin{tabular}{lrr}  
\toprule
Type &  ERBP L1 & ERBP L2 \\
\midrule
Adam   & 100 (0.00)  & 100 (0.00) \\
SGD     & 98 (0.06) & 96 (0.05) \\
\bottomrule

\end{tabular}

\end{table}

\textbf{Optimiser}:
We used both Stochastic Gradient Descent (SGD) and the Adam optimizer \citep{adam_ref} for training the ERBP L1 and L2. 
We observed faster convergence and greater improvement in the overall accuracy with the Adam compared to SGD. 
We observed similar results for both ERBP L1 and L2. 
Table \ref{tab:opt1} below summarizes the results of identity learning for both the optimizers with the regularization parameter $\lambda$ set to 1.
We observe that the SGD  does not reach full  generalization in this  setting, however it does  so at higher values of  $\lambda$.

\begin{figure}[htb]
 \centerline{\includegraphics[width=7cm]
 {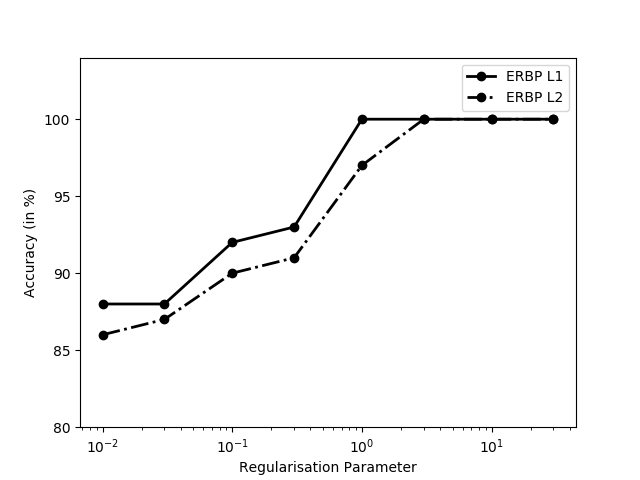}}
\caption{Test set classification accuracy (in \%) of the network with ERBP L1 and L2 when varying the regularization parameter $\lambda$ (shown in logarithmic scale) for identity learning with ($n=3$) and using the Adam optimizer.
}
 \label{fig:example3}
 
\end{figure}

\textbf{Regularization Factor}:
We varied the regularization factor $\lambda$ in the loss function of the ERBP models.  
We observed that a factor of 3 or above reliably leads to perfect generalization in identity learning task. 
Figure \ref{fig:example3}, shows how the effect depends on the size the regularization factor $\lambda$ using L1 and L2 loss functions for learning identity  relations.

\subsection{Jointly learning identity relations and bit patterns}

In order to test, whether the RBP and ERBP impedes other learning tasks, we also tested learning with some non-relational patterns that are based on values of specific neurons. 
The simplest form is that we have one bit that determines the target class. 
This case is learned with 100\% generalization performance for a network with 2 additional outputs for the pattern classification. 

Furthermore, we tested the learning of classification based on all even or odd elements in the vector being $0$. 
In this case, we also get perfect generalization for $\lambda$ up to 10. 
For $\lambda = 30$ we see first deterioration of the accuracy, but there is a wide range of values where both the identity relations and the even/odd classification generalize perfectly. 

\section{Conclusions}
Identity based relations are a fundamental form of relational learning. 
In this work, we re-visit the problem of learning identity relations using standard neural networks and show that creating a weight prior on the network weights leads to generalisable solutions of learning identity based relations. 
This also did not affect learning of non-relational patterns in our preliminary experiments, although more thorough testing remains to be done here.
We believe that addressing these issues and coming up with effective solutions is necessary for more higher level relational learning tasks and also is relevant to address problems in general neural network learning. 
In future, we would like to  extend this work towards learning other complex relational learning tasks and come with effective ways of creating an inductive bias using weight priors within the standard neural network architectures.

\bibliography{neurips_2019}

\end{document}